\documentclass[10pt,twocolumn,letterpaper]{article}

\usepackage{cvpr}
\usepackage{times}
\usepackage{epsfig}
\usepackage{graphicx}
\usepackage{amsmath}
\usepackage{amssymb}

\usepackage{ifthen}
\usepackage{color}
\usepackage{bigstrut}
 \usepackage{stfloats}
\usepackage[table]{xcolor}
\usepackage{multirow}
\usepackage[utf8x]{inputenc} 
\usepackage{tabularx}
\usepackage[normalem]{ulem}
\newcommand{\myfootnote}[2]{{%
  \let\thempfn\relax
  \footnotetext[0]{$^#1$\emph{#2}}
}}
\usepackage[pagebackref=true,breaklinks=true,letterpaper=true,colorlinks,citecolor=blue,linkcolor=blue,bookmarks=false]{hyperref}

\definecolor{frenchblue}{rgb}{0.0, 0.45, 0.73}
\definecolor{gray}{rgb}{0.5,0.5,0.5} 
\definecolor{green}{rgb}{0, 0.4, 0} 
\definecolor{orange}{rgb}{1, 0.5, 0} 	
\definecolor{mahogany}{rgb}{0.75, 0.25, 0.0}
\definecolor{purple}{rgb}{0.6, 0, 0.6}
\definecolor{darkgreen}{rgb}{0, 0.4, 0.4} 
\definecolor{teal}{rgb}{0.0, 0.5, 0.5}
\definecolor{aaaa}{rgb}{0.55, 0.1, 0.7}
\definecolor{red}{rgb}{1.0, 0, 0}

\newboolean{revising}
\setboolean{revising}{true}
\ifthenelse{\boolean{revising}}
{

} {

}

\cvprfinalcopy 

\def\cvprPaperID{1690} 

\begin{document}

\title{LayoutMP3D: Layout Annotation of Matterport3D}

\author{
    Fu-En Wang$^{* 1}$ \\
    {\tt\small fulton84717@gapp.nthu.edu.tw}
    \and
    Yu-Hsuan Yeh$^{* 2}$\\
    {\tt\small yuhsuan.eic08g@nctu.edu.tw}
    \and
    Min Sun$^{1}$\\
    {\tt\small sunmin@ee.nthu.edu.tw}
    \and
    Wei-Chen Chiu$^{2}$\\
    {\tt\small walon@cs.nctu.edu.tw}
    \and
    Yi-Hsuan Tsai$^{3}$\\
    {\tt\small wasidennis@gmail.com} 
}

\maketitle

\begin{abstract}
    Inferring the information of 3D layout from a single equirectangular panorama is crucial for numerous applications of virtual reality or robotics (e.g., scene understanding and navigation). To achieve this, several datasets are collected for the task of 360$^\circ$ layout estimation.
    To facilitate the learning algorithms for autonomous systems in indoor scenarios, we consider the Matterport3D dataset with their originally provided depth map ground truths and further release our annotations for layout ground truths from a subset of Matterport3D.
    %
    %
    As Matterport3D contains accurate depth ground truths from time-of-flight (ToF) sensors, our dataset provides both the layout and depth information, which enables the opportunity to explore the environment by integrating both cues. Our annotations are made available to the public at \url{https://github.com/fuenwang/LayoutMP3D}.
\end{abstract}

\myfootnote{1}{National Tsing Hua University}
\myfootnote{2}{National Chiao Tung University }
\myfootnote{3}{NEC Labs America}
\myfootnote{*}{The authors contribute equally to this paper.}

\section{Introduction}
As 360$^\circ$ cameras become popular in recent years, many approaches of applying deep neural network to panorama are proposed. Due to the large field of view (FoV), 360$^\circ$ cameras meet the requirements of sensing the environment efficiently for autonomous systems.  To achieve this, several approaches of scene understanding for 360$^\circ$ images are also proposed. For example, OmniDepth~\cite{omnidepth} focuses on the task of monocular 360$^\circ$ depth estimation. However, only using depth information is usually not effective enough for autonomous systems as the depth cannot provide a overall understanding of the indoor environment. To solve this problem, layout information can be utilized because the layout provides a rough geometry of rooms, which makes autonomous systems have high-level perception of the surroundings. Hence, the task of 360$^\circ$ layout estimation has also been studied these years.

As both the information of depth and layout are crucial to autonomous systems, we can imagine that the consideration of using both depth and layout would become more important in the near future. However, to the best of our knowledge, this direction has not been widely explored.
One of the key reasons is that there is no existing dataset which provides both depth and layout ground truths. Thus, to open more opportunities for this topic, we release the annotations of layout ground truths based on the subset of Matterport3D~\cite{mp3d}, which already contains accurate depth ground truths from time-of-flight (ToF) sensors.

\section{Related Work}
\paragraph{360$^\circ$ Perception.} As a single panorama needs to contain 360$^\circ$ information, the common formats of perspective image with normal FoV cannot be directly used. Therefore, the equirectangular and cubemap projections are commonly adopted to store a single panorama. For equirectangular projection, each pixel is mapped to a certain longitude and latitude which can be converted to a point on a unit sphere. However, using equirectangular projection would introduce severe distortions, in which standard approaches of deep neural networks cannot be directly applied. As a result, several convolutional methods designed for equirectangular projection are introduced ~\cite{sphericalcnns, Esteves_eccv18, KTN, yu-sphericalcnn}.

Although the above-mentioned methods are aware of the distortion on the equirectangular projection, the computational cost is also large. Hence, the extension to 360$^\circ$ images for conventional convolution has been also studied. Cheng~\etal~\cite{cubepadding} first converts an equirectangular image into cubemap projection. In this way, each face of cubemap can be viewed as a perspective image. As the faces in a cube are connected with the neighboring ones, the cube padding technique is proposed to make conventional convolutional operations aware of these connections.

\paragraph{360{\boldmath$^\circ$} Depth Estimation.} Zioulis~\etal~\cite{omnidepth} first propose OmniDepth for monocular 360$^\circ$ depth estimation. Based on the spherical convolution~\cite{yu-sphericalcnn}, they use UResNet and ResNet to predict 360$^\circ$ depth on equirectangular projection. Moreover, the dataset ``360D'' is also collected to verify their proposed method. With cube padding~\cite{cubepadding}, Wang~\etal~\cite{ouraccv} first propose the self-supervised training scheme for 360$^\circ$ depth estimation and the dataset ``PanoSUNCG'' is also collected. By using MINOS~\cite{minos} to collect data in reconstructed scenes of Matterport3D~\cite{mp3d} and Stanford2D3D~\cite{stanford2d3d}, Wang~\etal~\cite{Wang_icra_2020} propose the 360SD-Net method for stereo 360$^\circ$ depth estimation.

\paragraph{360{\boldmath$^\circ$} Layout Estimation.} Early efforts have been made via decomposing the equirectangular image into several perspective images and combining their information to recover the room layout~\cite{panocontext, ceilingview, automatic}. With the supervision of layout boundary and corners, Zou~\etal~\cite{layoutnet} propose LayoutNet with an end-to-end training scheme for layout estimation. Sun~\etal~\cite{horizonnet} propose HorizonNet by applying bi-directional LSTM to smooth the layout boundary along the horizontal direction. With the fusion between ceiling perspective and panorama views, Yang~\etal~\cite{dula-net} propose DuLa-Net to estimate the floorplan and the corresponding height to recover 3D layouts from 2D semantic maps. To verify their proposed method for general cases of layout (e.g., complex cases other than cuboid and L-type rooms), the dataset ``Realtor360'' is collected, which currently is the largest dataset containing diverse scenarios for layout estimation. 
\section{Data Collection}

\begin{figure}[h]
    \centering
    \includegraphics[width=0.85\columnwidth]{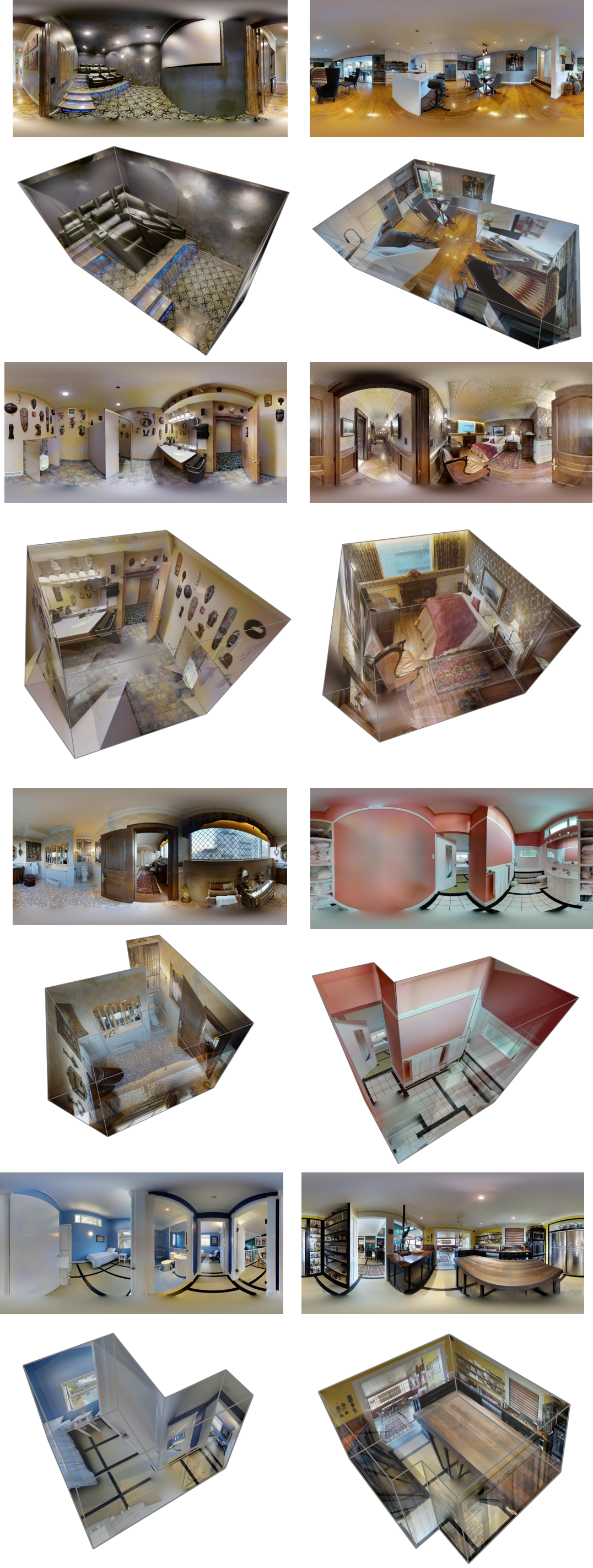}
    \caption{Example layouts of our dataset.}
    \label{fig:examples}
\end{figure}

For training a deep neural network for depth and layout estimation, a high-quality dataset with large amount of images and diverse cases of layout is essential. However, the existing largest dataset of layout estimation (i.e., Realtor360) only provides layout ground truths without depth maps from the laser scanner. On the other hand, Matteport3D~\cite{mp3d} only provides depth ground truths without the layout information. As the collection of depth ground truth is more expensive than the layout one, we manually annotate layout ground truths of Matterport3D. For our labeling, we use the annotation tool from DuLa-Net~\cite{dula-net} to label Manhattan cases and ignore non-Manhattan ones. Table~\ref{tab:statics} shows the statistics of our dataset. In total, our dataset contains 2285 (1830 for training, 455 for testing) panoramas with both depth and layout ground truths. We also provide some example layouts in Figure \ref{fig:examples}.

\begin{table}[htbp]
  \centering
  \small
  \caption{The statistics of our dataset.}
  \vspace{1mm}
    \begin{tabular}{|c|c|c|c|c|}
    \hline
    4 corners & 6 corners & 8 corners & 10+ corners & Total \bigstrut\\
    \hline
    1243  & 515   & 308   & 219   & 2285 \bigstrut\\
    \hline
    \end{tabular}%
  \label{tab:statics}%
\end{table}%

\section{Conclusions}
In this paper, we release the layout annotations of Matterport3D which contains comparable amount with Realtor360. As Matterport3D already provides depth ground truths, our annotation could open more opportunities for the direction of jointly considering the depth and layout information.
In this way, the navigation and scene understanding tasks of indoor environment could be benefited from the dataset we provide.

{\small
\bibliographystyle{ieee_fullname}

\begin{thebibliography}{10}\itemsep=-1pt

\bibitem{stanford2d3d}
Iro Armeni, Sasha Sax, Amir~Roshan Zamir, and Silvio Savarese.
\newblock Joint 2d-3d-semantic data for indoor scene understanding.
\newblock {\em CoRR}, 2017.

\bibitem{mp3d}
Angel Chang, Angela Dai, Thomas Funkhouser, Maciej Halber, Matthias Niessner,
  Manolis Savva, Shuran Song, Andy Zeng, and Yinda Zhang.
\newblock Matterport3{D}: Learning from rgb-d data in indoor environments.
\newblock 2017.

\bibitem{cubepadding}
Hsien-Tzu Cheng, Chun-Hung Chao, Jin-Dong Dong, Hao-Kai Wen, Tyng-Luh Liu, and
  Min Sun.
\newblock Cube padding for weakly-supervised saliency prediction in 360°
  videos.
\newblock In {\em IEEE Conference on Computer Vision and Pattern Recognition
  (CVPR)}, 2018.

\bibitem{sphericalcnns}
Taco~S. Cohen, Mario Geiger, Jonas Köhler, and Max Welling.
\newblock Spherical {CNN}s.
\newblock In {\em International Conference on Learning Representations (ICLR)},
  2018.

\bibitem{Esteves_eccv18}
Carlos Esteves, Christine Allen-Blanchette, Ameesh Makadia, and Kostas
  Daniilidis.
\newblock Learning so(3) equivariant representations with spherical cnns.
\newblock In {\em European Conference on Computer Vision (ECCV)}, 2018.

\bibitem{ceilingview}
G. Pintore, V. Garro, F. Ganovelli, E. Gobbetti, and M. Agus.
\newblock Omnidirectional image capture on mobile devices for fast automatic
  generation of 2.5d indoor maps.
\newblock In {\em IEEE Winter Conference on Applications of Computer Vision
  (WACV)}, 2016.

\bibitem{minos}
Manolis Savva, Angel~X. Chang, Alexey Dosovitskiy, Thomas Funkhouser, and
  Vladlen Koltun.
\newblock {MINOS}: Multimodal indoor simulator for navigation in complex
  environments.
\newblock {\em CoRR}, 2017.

\bibitem{KTN}
Yu{-}Chuan Su and Kristen Grauman.
\newblock Kernel transformer networks for compact spherical convolution.
\newblock {\em CoRR}, 2018.

\bibitem{yu-sphericalcnn}
Yu-Chuan Su and Kristen Grauman.
\newblock Learning spherical convolution for fast features from 360\textdegree
  imagery.
\newblock In I. Guyon, U.~V. Luxburg, S. Bengio, H. Wallach, R. Fergus, S.
  Vishwanathan, and R. Garnett, editors, {\em Advances in Neural Information
  Processing Systems (NIPS)}. 2017.

\bibitem{horizonnet}
Cheng Sun, Chi-Wei Hsiao, Min Sun, and Hwann-Tzong Chen.
\newblock Horizonnet: Learning room layout with 1d representation and pano
  stretch data augmentation.
\newblock In {\em IEEE Conference on Computer Vision and Pattern Recognition
  (CVPR)}, 2019.

\bibitem{ouraccv}
Fu-En Wang, Hou-Ning Hu, Hsien-Tzu Cheng, Juan-Ting Lin, Shang-Ta Yang, Meng-Li
  Shih, Hung-Kuo Chu, and Min Sun.
\newblock Self-supervised learning of depth and camera motion from 360$^\circ$
  videos.
\newblock In {\em Asian Conference on Computer Vision (ACCV)}, 2018.

\bibitem{Wang_icra_2020}
Ning-Hsu Wang, Bolivar Solarte, Yi-Hsuan Tsai, Wei-Chen Chiu, and Min Sun.
\newblock 360sd-net: 360$^\circ$ stereo depth estimation with learnable cost
  volume.
\newblock {\em arXiv:1911.04460}, 2019.

\bibitem{dula-net}
Shang-Ta Yang, Fu-En Wang, Chi-Han Peng, Peter Wonka, Min Sun, and Hung-Kuo
  Chu.
\newblock Dula-net: {A} dual-projection network for estimating room layouts
  from a single {RGB} panorama.
\newblock In {\em IEEE Conference on Computer Vision and Pattern Recognition
  (CVPR)}, 2019.

\bibitem{automatic}
Yang Yang, Shi Jin, Ruiyang Liu, Sing Bing~Kang, and Jingyi Yu.
\newblock Automatic 3d indoor scene modeling from single panorama.
\newblock In {\em IEEE Conference on Computer Vision and Pattern Recognition
  (CVPR)}, 2018.

\bibitem{panocontext}
Yinda Zhang, Shuran Song, Ping Tan, and Jianxiong Xiao.
\newblock Panocontext: A whole-room 3d context model for panoramic scene
  understanding.
\newblock In {\em European Conference on Computer Vision (ECCV)}, 2014.

\bibitem{omnidepth}
Nikolaos Zioulis, Antonis Karakottas, Dimitrios Zarpalas, and Petros Daras.
\newblock Omnidepth: Dense depth estimation for indoors spherical panoramas.
\newblock In {\em European Conference on Computer Vision (ECCV)}, 2018.

\bibitem{layoutnet}
Chuhang Zou, Alex Colburn, Qi Shan, and Derek Hoiem.
\newblock Layoutnet: Reconstructing the 3d room layout from a single rgb image.
\newblock In {\em IEEE Conference on Computer Vision and Pattern Recognition
  (CVPR)}, 2018.

\end{thebibliography}

}

\end{document}